\documentclass[10pt,twocolumn,letterpaper]{article}

\usepackage{cvpr}
\usepackage{times}
\usepackage{epsfig}
\usepackage{graphicx}
\usepackage{amsmath}
\usepackage{amssymb}
\usepackage{caption}
\usepackage{subcaption}
\usepackage{comment}

\newcommand{\minisection}[1]{\vspace{0.04in} \noindent {\bf #1}\ \ }

\usepackage{multirow}

\usepackage[pagebackref=true,breaklinks=true,letterpaper=true,colorlinks,bookmarks=false]{hyperref}

\cvprfinalcopy

\ifcvprfinal\fi
\begin{document}

\title{Orderless Recurrent Models for Multi-label Classification}

\author{Vacit Oguz Yazici$^{1,2}$, Abel Gonzalez-Garcia$^{1}$, Arnau Ramisa$^{2,3}$, \\ Bartlomiej Twardowski$^{1}$, Joost van de Weijer$^{1}$\\
$^{1}$ Computer Vision Center, Universitat Autonoma de Barcelona, Barcelona, Spain \\
$^{2}$ Wide-Eyes Technologies, Barcelona, Spain $^{3}$ Universitat de Vic, Barcelona, Spain\\
{\tt\small \{voyazici,agonzalez,bartlomiej.twardowski,joost\}@cvc.uab.es, arnau.ramisa@uvic.cat}
}

\maketitle

\begin{abstract}
   Recurrent neural networks (RNN) are popular for many computer vision tasks, including multi-label classification. Since RNNs produce sequential outputs, labels need to be ordered for the multi-label classification task. Current approaches sort labels according to their frequency, typically ordering them in either rare-first or frequent-first. These imposed orderings do not take into account that the natural order to generate the labels can change for each image, e.g.\ first the dominant object before summing up the smaller objects in the image. Therefore, we propose ways to dynamically order the ground truth labels with the predicted label sequence. This allows for faster training of more optimal LSTM models.
   Analysis evidences that our method does not suffer from duplicate generation, something which is common for other models. Furthermore, it outperforms other CNN-RNN models, and we show that a standard architecture of an image encoder and language decoder trained with our proposed loss obtains the state-of-the-art results on the challenging MS-COCO, WIDER Attribute and PA-100K and competitive results on NUS-WIDE.
\end{abstract}

\section{Introduction}

RNNs have demonstrated good performance in many tasks that require processing variable length sequential data. One of the most popular types of RNN is the Long-Short Term Memory networks (LSTM)~\cite{hochreiter1997long}. LSTMs improve over earlier RNNs, especially addressing the vanishing gradient problem, and have advanced the state of the art in machine translation~\cite{sutskever2014sequence} and speech recognition~\cite{graves2013hybrid}, among other tasks. They have also been combined with deep Convolutional Neural Networks (CNN) and used for computer vision tasks, such as image captioning~\cite{vinyals2015show}, and video representations~\cite{srivastava2015unsupervised}. Furthermore, LSTMs have been shown to be useful for traditionally non-sequential tasks, like multi-label classification~\cite{chen2018order,jin2016annotation,liu2017semantic,wang2016cnn}.

\begin{figure}
    \includegraphics[width=\linewidth]{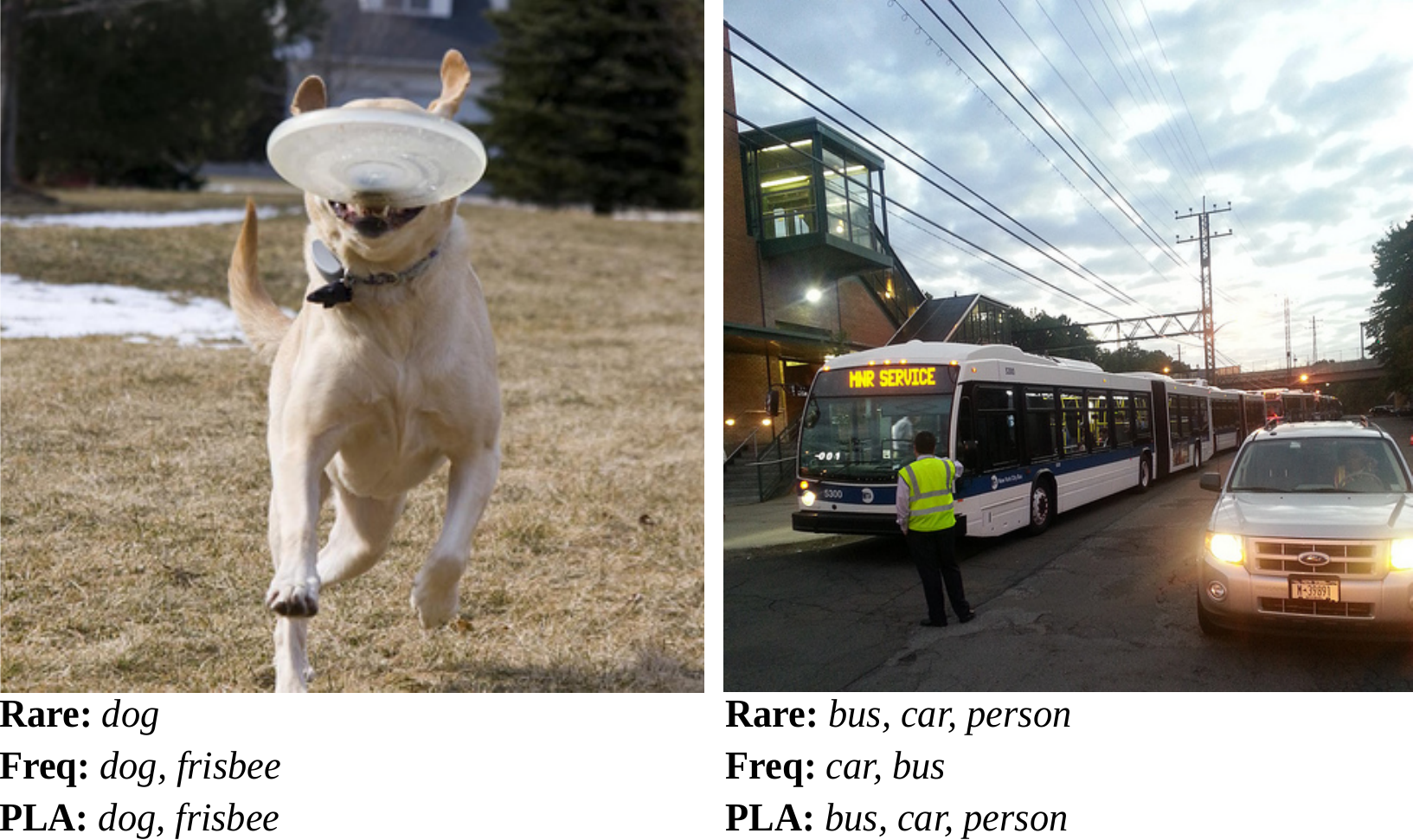}
    \caption{\small
    Estimated labels for various approaches. In the \textbf{Rare} (rare-first) approach bigger and more frequent classes might cause other classes to be ignored (\emph{frisbee} in the left figure), meanwhile in the \textbf{Freq} (frequent-first) approach smaller frequent classes are ignored (\emph{person} in the right figure). Our approach \textbf{PLA} circumvents these problems and correctly assigns the labels to both images.} 
    \label{fig:freq_rare_problems}
\end{figure}

Multi-label classification is the task of assigning a wide range of visual concepts to images. These concepts could include object classes or actions, but also attributes such as colors, textures, materials, or even more abstract notions like mood. The large variety of concepts makes this a very challenging task and, to successfully address it, methods should learn the dependencies between the many concepts: boats are not common in office spaces, and penguins are seldom seen in deserts.
Another problem of multi-label classification is the fact that similarities between classes may make the model uncertain about a particular object (e.g. it could be either a bicycle or a motorcycle) while being sure that both are not simultaneously present in the image. Consequently, it should choose one of the labels and not both, but traditional approaches like Binary Cross-Entropy (BCE) do not discount evidence already used in support of another label, and would predict both.  In practice, these dependencies between the labels turn the task of multi-label classification into a structured labelling problem~\cite{liu2017semantic}.

Image captioning, where the task is to generate a natural language sentence describing the image, is highly related to multi-label classification. The main difference is that in image captioning the ordering constraint imposed by the recurrent neural network comes naturally, as sentences have a sequential nature, and RNNs are considered the appropriate model to generate an ordered list of words~\cite{vinyals2015show, xu2015show}. Recently, it was found that recurrent networks also obtain good results in (orderless) structured labelling tasks like multi-label classification, and that they were good at modelling the dependencies in label space. Typically this is implemented by replacing the BCE ``multi-head" of the network with an LSTM module, using it to generate a variable length sequence of labels, plus a termination token. 
However, this approach has a caveat: the LSTM loss will penalize otherwise correct predictions if they are not generated in the same ordering as in the ground truth label sequence. This seriously hinders convergence, complicates the training process, and generally results in inferior models.

Several recent works have tried to address this issue by imposing an arbitrary, but consistent, ordering to the ground truth label sequences~\cite{wang2016cnn, jin2016annotation}. The rationale is that if the labels are presented always in the same order, the network will, in turn, predict them in the same order as well.
Despite alleviating the problem, these approaches are short of solving it, and many of the original issues are still present. For example, in an image that features a clearly visible and prominent \textit{dog}, the LSTM may chose to predict that label first, as the evidence for it is very large. However, if \textit{dog} is not the label that happened to be first in the chosen ordering, the network will be penalized for that output, and then penalized again for not predicting \textit{dog} in the ``correct" step according to the ground truth sequence.
In this paper we observe that this leads to more difficult convergence, as well as sub-optimal results, like a label being predicted several times for the same image by the trained model.

In contrast with related works, we do not impose a predefined order to the output sequence, since this does not respond to any real constraint that the model should fulfill. Instead, we dynamically chose the ordering that minimizes the loss during training by re-arranging the ground truth sequence to match as closely as possible the sequence of predicted labels. We propose two ways of doing that: \emph{predicted label alignment} (PLA) and \emph{minimal loss alignment} (MLA). We empirically show that these approaches lead to faster training (see Figure~\ref{fig:losses}), and also eliminate other nuisances like repeated labels in the predicted sequence. Furthermore, we obtain state-of-the-art results on the MS-COCO, WIDER Attribute and PA-100K datasets.

\section{Related Work}
\label{sec:related_work}
\minisection{Deep Recurrent Networks} 
Recurrent neural networks~\cite{RumelhartHintonWIlliams1986}, are neural networks that include loops, and can process the same input (plus an internal state used for passing messages between iterations) several times with the same weights. The original formulation of RNNs is notoriously difficult to train because of the exploding/vanishing gradient problems, which are exacerbated with long input sequences. Later research found solutions to these problems, including the particularly successful models Gated Recurrent Unit~\cite{cho2014learning} and Long-Short Term Memory~\cite{hochreiter1997long}.

Despite originally being designed for sequential data, LSTM networks have also been used for orderless data, or sets~\cite{vinyals2015order, chen2018order}. Vinyals et al.~\cite{vinyals2015order} explores the different types of orderless data that can be processed by an LSTM network, and proposes different architectures and training procedures to deal with them. Chen et al.~\cite{chen2018order} proposes a method for order-free usage of recurrent networks for multi-label image annotation. Both these methods are discussed in more detail after we have introduced our approach to the training of orderless recurrent networks (see Section \ref{sec:orderless}).

\minisection{Multi-label classification}
Unlike in traditional (single-label) classification, in multi-label classification each image can be associated with more than one concept. Yet, initial approaches for multi-label classification in the literature treat each occurrence of a label independently from the others~\cite{wei2014cnn, gong2013deep}, thus not taking advantage of label correlations.

Earlier works that tried to leverage label correlations exploited graphical models such as Conditional Random Fields (CRFs)~\cite{ghamrawi2005collective} or Dependency Networks~\cite{guo2011multi}. Chen et al.~\cite{chen2015learning} combine CRFs with deep learning algorithms to explore dependencies between the output variables. Read et al.~\cite{read2011classifier} propose using a chain of binary classifiers to do multi-label classification. Most of the approaches mentioned come with relatively high computation costs since they need to model explicitly the pairwise label correlations.

On the other hand, RNN-based multi-label classification does not incur these high computation costs, since the low dimensional RNN layers work well to model the label correlations~\cite{wang2016cnn,jin2016annotation}. The idea to exploit RNN models to capture label correlations was originally proposed in~\cite{jin2016annotation} and~\cite{wang2016cnn}. Wang et al.~\cite{wang2016cnn} combine CNN and RNN architectures, and learn a joint image-label embedding space to characterize the label semantic dependencies. Since LSTMs produce sequential outputs, they use a frequent-first ordering approach. Jin et al.~\cite{jin2016annotation} use a CNN to encode images and input them to an RNN that generates the predictions. They use frequent-first, dictionary-order, rare-first and random order in their experiments and compare the results of different methods. Liu et al.~\cite{liu2017semantic} use a similar architecture, but they make the CNN and RNN models explicitly address the label prediction and label correlation tasks respectively. Instead of using a fully connected layer between the CNN and RNN models, they input class probabilities predicted by the CNN model to the RNN. In that way, they supervise both models during the training. They use rare-first ordering in their model to assign more importance to the less common labels. Chen et al.~\cite{chen2018order}  use a BCE loss to compute predictions in each time step to remove the order of the labels. However, none of these approaches adapt the order dynamically according to the predictions, and they only achieve marginal improvements over CNN models.
Concurrent to our work, Pineda et al.~\cite{pineda2019elucidating} present a comprehensive comparison between CNN and CNN-RNN methods on various datasets with different characteristics.

\minisection{Image captioning}
Earlier works on image captioning with deep learning have adapted the \emph{encoder-decoder} framework in which an RNN (usually an LSTM) is used to ``translate" image features into a sentence, one word at a time~\cite{ChenCVPR2015, DonahueCVPR2015, FangCVPR2015, JohnsonICCV2015, karpathy15cvpr, KirosTACL2015, MaoICLR2015, VinyalsCVPR2015, XuICML2015}. These image features are usually generated using a CNN that \textit{encodes} (or \textit{translates}) the image into a higher-level representation, and next an RNN \textit{decodes} this representation back into natural language. An important part of the success of this type of models is that the whole system is trained end-to-end, and so both components can co-adapt to yield the best results. See~\cite{Bai2018, Hossain2019} for recent surveys on image caption generation.

\vspace{-0.05cm}
\section{Method}
\begin{figure}
  \includegraphics[width=\linewidth]{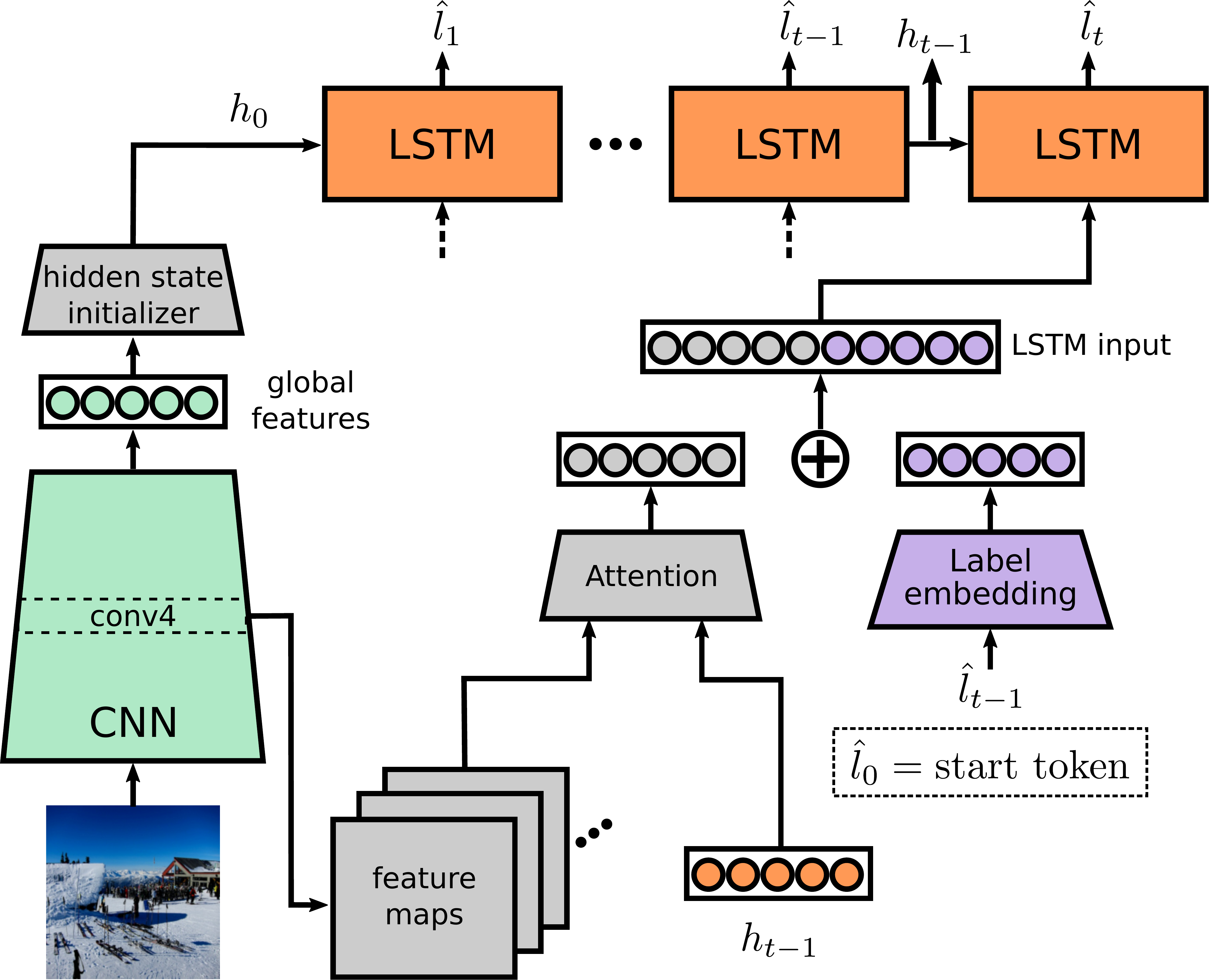}
  \caption{\small CNN-RNN architecture used in this paper, containing of an image CNN encoder, an LSTM text decoder and an attention mechanism. We show that this simple architecture can obtain state-of-the-art results by  substituting the loss function by an orderless loss function. }
  \label{fig:architecture}
  \vspace{-0.4cm}
\end{figure}
\subsection{Image-to-sequence model}
\vspace{-0.05cm}
For the task of multi-label classification we consider a CNN-RNN architecture, first proposed in \cite{wang2016cnn}. This type of model consists of a CNN (encoder) part that extracts a compact visual representation from the image, and of an RNN (decoder) part that uses the encoding to generate a sequence of labels, modeling the label dependencies. Different authors experimented with different choices of visual representation to feed to the RNN: in~\cite{wang2016cnn}, images and labels are projected to the same low-dimensional space to model the image-text relationship, while~\cite{liu2017semantic} uses the predicted class probabilities, and~\cite{jin2016annotation} experiments with different internal layers of the CNN. In our approach, we use the final fully connected layer to initialize the hidden state of the RNN. Once initialized, the RNN model predicts a new label every time step until an end signal is generated.

The choice of RNN typically used in CNN-RNN models is the Long-Short Term Memory. Unlike prior RNN models, LSTM mitigates the vanishing gradient problem by introducing a forget gate $f$, an input gate $i$ and an output gate $o$ to an RNN layer. With these gates, it can learn long term dependencies in a sequential input. The equations that govern the forward propagation through the LSTM at time step $t$ and with an input vector $x_t$ are the following:
\vspace{-0.1cm}
\begin{equation}
\begin{aligned}
f_t &= \sigma(W_fx_t + U_fh_{t-1} + b_f)\\
i_t &= \sigma(W_ix_t + U_ih_{t-1} + b_i)\\
o_t &= \sigma(W_0x_t + U_0h_{t-1} + b_0)\\
c_t &= f_t \odot c_{t-1} + i_t \odot tanh(W_cx_t + U_ch_{t-1} + bc) \\
ht &= o_t \odot tanh(c_t)
\end{aligned}
\vspace{-0.1cm}
\end{equation}
where $c_t$ and $h_t$ are the model cell and hidden states, while $i_t$, $f_t$, $o_t$ are the input, forget and output gates' activations respectively. $W$, $U$ and $b$ are the weights and biases to be learned, and the $\sigma$ and $tanh$ are the sigmoid and hyperbolic tangent functions respectively. At time step $t$, the model uses as input the predicted output embedding from the previous time step. The predictions for the current time step $t$ are computed in the following way:
\vspace{-0.1cm}
\begin{equation}
\begin{aligned}
x_t &= E \cdot {\hat{l}_{t-1}} \\
h_t &= \text{LSTM}(x_t, h_{t-1}, c_{t-1}) \\
p_t &= W \cdot h_t + b
\end{aligned}
\vspace{-0.1cm}
\end{equation}
where $E$ is a word embedding matrix and $\hat{l}_{t-1}$ is the predicted label index in the previous time step. The prediction vector is denoted by $p_t$, and $W$ and $b$ are the weights and the bias of the fully connected layer.

We also include the attention module that was proposed in~\cite{xu2015show}. Linearized activations from the fourth convolutional layer are used as input for the attention module, along with the hidden state of the LSTM at each time step, thus the attention module focuses on different parts of the image every time. These attention weighted features are then concatenated with the word embedding of the class predicted in the previous time step, and given to the LSTM as input for the current time step. As pointed out in~\cite{wang2016cnn}, it is hard to represent small objects with global features, so an attention module alleviates the problem of ignoring smaller objects during the prediction step. A diagram of our model architecture is provided in Figure~\ref{fig:architecture}. 
\subsection{Training recurrent models}
To train the model a dataset with pairs of images and sets of labels is used. Let $\left(I, L\right)$ be one of the pairs containing an image $I$ and its $n$ labels $L=\{l_1,l_2,...,l_{n}\}, \; l_i \in \mathbb{L}$, with $\mathbb{L}$ the set of all labels with cardinality $m=|\mathbb{L}|$, including the start and end tokens. 

\begin{figure}
  \begin{center}
    \includegraphics[width=\linewidth]{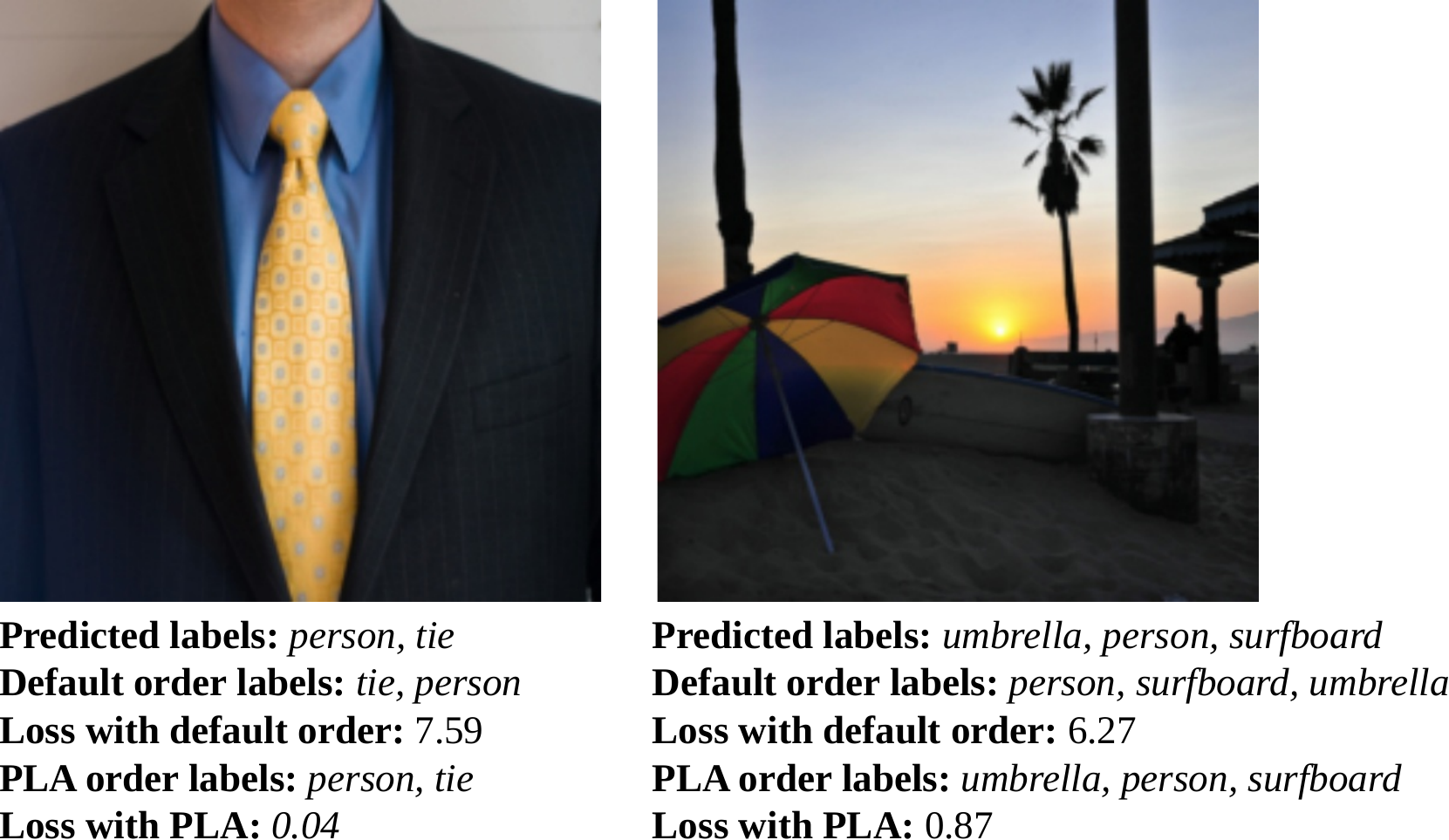}
    \caption{\small 
     Comparison of an ordered loss to our orderless PLA loss. Imposing any order (default order in this example) leads to high losses even though the labels are correct. PLA solves this problem by dynamically adapting the order.
   }
    \label{fig:loss_vis}
  \end{center}
\end{figure}

The predictions $p_t$ of the LSTM are collected in the matrix $P=[p_1\;p_2\;...\;p_n]$, with $P \in \mathbb{R}^{m\times n}$. When the number of predicted labels $k$ is larger than $n$, we only select the first $n$ prediction vectors. In case $k$ is smaller than $n$ we pad the matrix with empty vectors to obtain the desired dimensionality. We can now define the standard cross-entropy loss for recurrent models as: 

\begin{equation}
\begin{aligned}
& \mathcal{L} \; = & & \hspace{-0.15cm} tr\left( T\; log \left(P\right)\right) \\
& \text{with} & & T_{tj}  = 1\;{\rm{if}}\;l_t  = j\\ 
& & & T_{tj}  = 0\;{\rm{otherwise}}
\end{aligned} \label{eq:loss}
\end{equation}
where $T \in	\mathbb{R}^{n\times m}$ contains the ground truth label for each time step\footnote{Here we consider that $l_1=\{1,...,m\}$ is the class-index.}. The loss is computed by comparing the prediction of the model at step $t$ with the corresponding label at the same step of the ground truth sequence.
As can be seen in Equation~\ref{eq:loss}, the order of the ground truth labels is critical to determine the loss a given prediction will receive (see Figure~\ref{fig:loss_vis}). For inherently orderless tasks like multi-label classification, where labels often come in random order, it becomes essential to minimize unnecessary penalization, and several approaches have been proposed in the literature. The most popular solution to improve the alignment between ground truth and predicted labels consists on defining an arbitrary criteria by which the labels will be sorted. Wang et al.~\cite{wang2016cnn} count occurrences of labels in the dataset and sort the labels according to their occurrence in descending order, and is consequently called the \textit{frequent-first} approach. Jin et al.~\cite{jin2016annotation} use a \textit{rare-first} approach and \textit{dictionary-order} in addition to the frequent-first approach. Unlike the frequent-first approach, the rare-first promotes the rare classes in the dataset, while dictionary-order sorts the labels in alphabetical order. The rare-first approach was also adopted by Liu et al.~\cite{liu2017semantic}.
Sorting the ground truth labels with a fixed, arbitrary, criteria is shown to improve results with respect to using a random ordering, since the network can learn to predict in the defined order, and avoid part of the loss. However, this will delay convergence, as the network will have to learn the arbitrary ordering in addition to predicting the correct labels given the image. Furthermore, any misalignment between the predictions and the labels will still result in higher loss and misleading updates to the network. Additionally, the frequency of a label in a dataset is independent of the size of the object in a given image. Less frequent but bigger objects can cause the LSTM prediction to stop earlier because of their dominance in the image and their ranking in the prediction step. This issue can be observed in Figure~\ref{fig:freq_rare_problems}, both for the frequent-first and rare-first approaches.

\subsection{Orderless recurrent models}\label{sec:orderless}
To alleviate the problems caused by imposing a fixed order to the labels, we propose to align them to the predictions of the network before computing the loss. We consider two different strategies to achieve this.

The first strategy, called \emph{minimal loss alignment (MLA)}  is computed with:
\begin{equation}
\begin{aligned}
 & \mathcal{L} = \mathop{\min}\limits_T & tr\left( {T\log \left( P \right)} \right) \\[-7pt]
 & \text{subject to} & T_{tj} \in \{0,1\}, & \sum\nolimits_j {T_{tj}  = 1},\\[-6pt]
 & & \sum\nolimits_t {T_{tj}  = 1} & \;\;\; \forall j \in L, \\[-5pt]
 & & \sum\nolimits_t {T_{tj}  = 0} & \;\;\; \forall j \notin L
\end{aligned}\label{eq:MLA}
\end{equation}
where $T \in \mathbb{R}^{n\times m}$ is a permutation matrix, which is constrained to have a ground truth label for each time step: $\sum_j T_{tj}  = 1$, and that each label in the ground truth $L$ should be assigned to a time step. The matrix $T$ is chosen in such a way as to minimize the summed cross entropy loss. This minimization problem is an assignment problem and can be solved with the Hungarian~\cite{kuhn1955hungarian} algorithm. 

We also consider the \emph{predicted label alignment (PLA)} solution. If we predict a label which is in the set of ground truth labels for the image, then we do not wish to change it. That leads to the following optimization problem:

\begin{equation}
\begin{aligned}
 & \mathcal{L} = \mathop{\min}\limits_T 
 & & tr\left( {T\log \left( P \right)} \right) \\[-7pt]
 & \text{subject to}
 & & T_{tj} \in \{ 0,1\}, \; \sum\nolimits_j {T_{tj}  = 1},\\[-6pt]
 & & & T_{tj} = 1\;\text{if} \;\; \hat{l}_t \in L \;\text{and} \; j=\hat{l}_t , \\[-4pt]
 & & & \sum\nolimits_t {T_{tj}  = 1}\;\;\forall\;j\in L,\\[-4pt]
 & & & \sum\nolimits_t {T_{tj}  = 0}\;\;\forall\;j\notin L
 \end{aligned}\label{eq:PLA}
\end{equation}
where $\hat{l}_t$ is the label predicted by the model at step $t$. Here we first fix those elements in the matrix $T$ for which we know that the prediction is in the ground truth set $L$, and apply the Hungarian algorithm to assign the remaining labels (with same constraints as Eq.~\ref{eq:MLA}). This second approach results in higher losses than the first one (Eq.~\ref{eq:MLA}), since there are more restrictions on matrix $T$. Nevertheless, this method is more consistent with the labels which were actually predicted by the LSTM.

\begin{figure}
  \includegraphics[width=\linewidth]{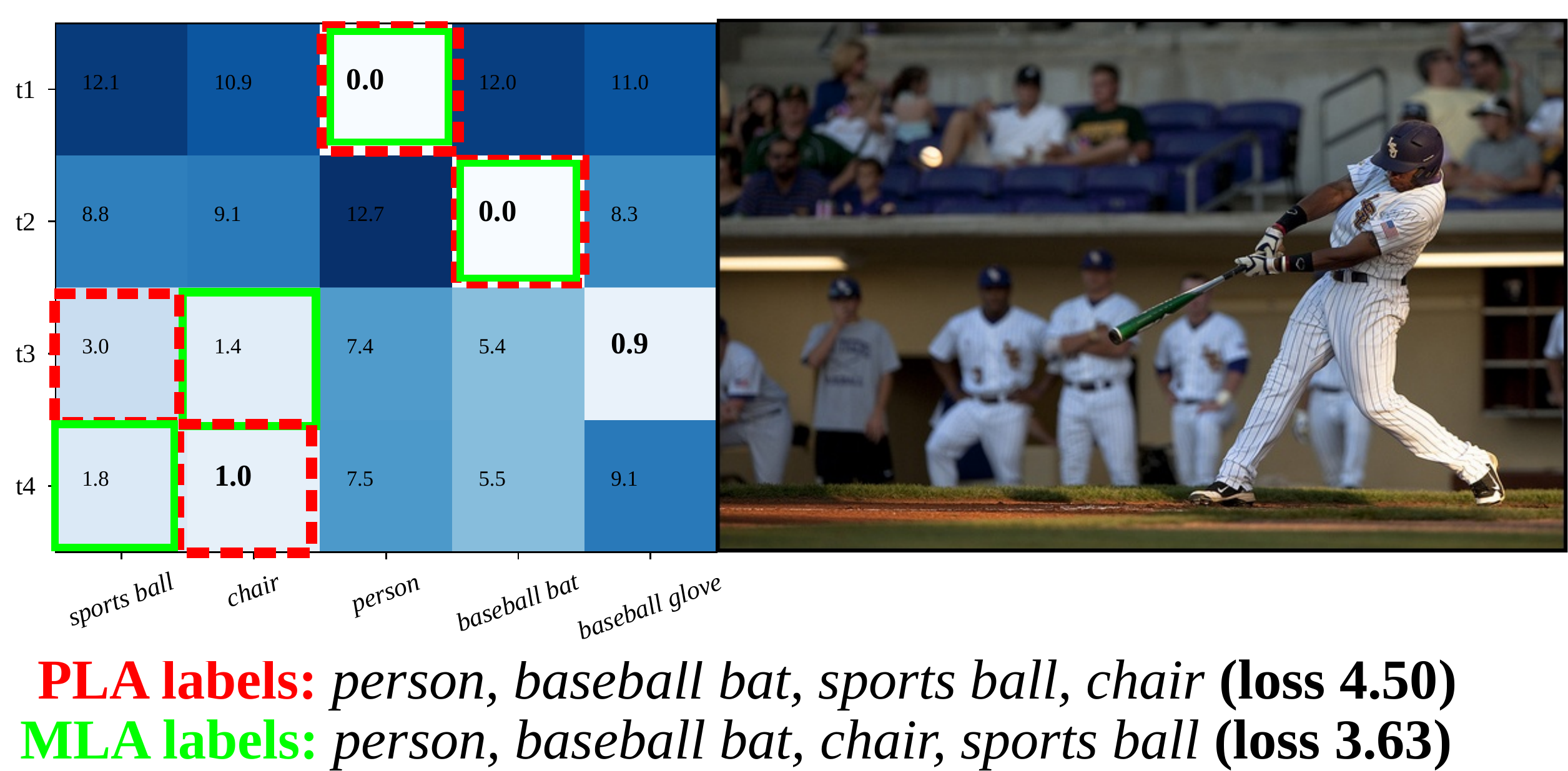}
  \caption{\small The cost matrix, image and different label orders decided by PLA and MLA (best viewed in color). Predicted classes are bolded. See text for explanation.}
  \label{fig:cost_matrix}
\vspace{-0.2cm}
\end{figure}

To further illustrate our proposed approach to train orderless recurrent models we consider an example image and its cost matrix (see Figure~\ref{fig:cost_matrix}). The cost matrix shows the cost of assigning each label to the different time steps. The cost is computed as the negative logarithm of the probability at the corresponding time step. Although the MLA approach achieves the order that yields the lowest loss, in some cases this can cause misguided gradients as it does in the example in the figure. The MLA approach puts the label \textit{chair} in the time step $t_3$, although the network already predicts it in the time step $t_4$. Therefore, the gradients force the network to output \textit{chair} instead of \textit{sports ball} although \textit{sports ball} is also one of the labels.

Orderless training of recurrent models has been previously addressed in~\cite{vinyals2015order, chen2018order}. Vinyals et al.~\cite{vinyals2015order} study the usage of recurrent models to represent sets for which no apparent order of the elements exists. Their method considers two phases: first a uniform prior over all orders is assumed for an initial number of iterations of training, after which in the second phase ancestral sampling is used to sample an ordering. Unlike our method, which proposes to adapt the label order according to the predicted order, their method aims to find the optimal order of the labels (without considering the predicted order). Their method has only been evaluated on a toy problem. More related to our work is the research of Chen et al.~\cite{chen2018order}, which applies a recurrent model without imposing any ordering. This is done by estimating all the labels in the image at every step of the recurrent model. They replace the standard cross entropy loss of the LSTM by a binary cross entropy (BCE). A drawback of this approach is that the LSTM will repeat labels already predicted before. Therefore an additional module needs to be introduced which prevents the method from repeating already predicted labels. An additional drawback of this method is that there is no end-token, so a threshold should be learned to stop the sequence. 

\section{Experiments}
\subsection{Datasets and setting}
We evaluate our models on four datasets: MS-COCO \cite{lin2014microsoft}, NUS-WIDE \cite{nus-wide-civr09}, WIDER Attribute \cite{li2016human} and PA-100K \cite{liu2017hydraplus}. \textbf{MS-COCO} is used for image segmentation, image captioning, and object detection. It can be also used for multi-label classification since it has labels for 80 objects. It consists of 82,081 training and 40,137 test images. \textbf{NUS-WIDE} consists of 269,648 images with a total number of 5,018 unique labels. However, annotations for 81 labels are more trustworthy and used for evaluation. After removing the images that do not belong to any of the 81 labels, 209,347 images remain. Following~\cite{johnson2015love, liu2017semantic}, we use 150,000 of these images for training, and rest for testing. For a fair comparison, we create 3 different splits and we pick the best scores on each split and average them to get the final scores. \textbf{WIDER Attribute} is a dataset which has 14 human attributes in 13,789 images with 57,524 annotated bounding boxes (28,345 for training and 29,179 for test). \textbf{PA-100K} is built for evaluating a pedestrian attribute recognition task. It consists of 100,000 pedestrian images with 26 attributes. The size of the training, validation and test sets are 80,000, 10,000 and 10,000 respectively.

\minisection{Evaluation metrics:}
We use \textit{per-class} and \textit{overall} precision, recall and F1 scores. The \textit{per-class} metric averages precision and recall scores for each class, and the geometric mean of these averaged values gives the \textit{per-class} F1 score. In the \textit{overall} metric, precision and recall scores are computed for all images, and the geometric mean of precision and recall gives the \textit{overall} F1 score. Only for PA-100K dataset, instead of evaluating accuracy of each label independently, we evaluate accuracy of image-wise class predictions to be able to compare the results with other models. Next, we are interested to see if our method actually adapts dynamically the order to the image, or just learns another (more optimal) fixed order of the classes. To this end, we use an \textit{order-rigidness} measure on the test set. For each pair of classes, two possible orderings exist (e.g.~for classes A and B that would be A-B or B-A); to compute the order-rigidness, we add the number of occurrences of the most frequent order for each pair of classes in the same image, and divide it by the total number of co-occurrences of any pair. We remove all but one of every duplicate prediction without penalization. We show order-rigidness and the percentage of images with duplicate predictions in Table~\ref{tab:analysis}.

\minisection{Network training:}
We implemented the architecture (see Figure~\ref{fig:architecture}) using the PyTorch framework~\cite{paszke2017automatic}. For the encoder part and BCE models we use the VGG16~\cite{simonyan2014very}, ResNet-50 (for PA-100K) and ResNet-101~\cite{he2016deep} architectures, and the decoder part is an LSTM with a 512 dimensional internal layer. The word embeddings learned during the training have dimension 256, and the attention module, 512. To train the BCE models, the SGD optimizer is used with learning rate $0.01$ and momentum $0.9$. For the LSTM models the encoder and the decoder are trained with the ADAM optimizer and Stochastic Weight Averaging~\cite{izmailov2018averaging} with a cyclical learning rate scheduler, decreasing from $10^{-3}$ to $10^{-6}$ in 3 iterations. The BCE models are trained for 40 epochs, and if no improvement is observed after 3 epochs, then we multiply the current learning rate by $0.1$. For the LSTM models, we fine-tune from the best BCE model and train for 30 epochs more. All the BCE models are pretrained on ImageNet~\cite{russakovsky2015imagenet}. Finally, we do not use the beam search algorithm; we just greedily take the maximum predicted output. Random affine transformations and contrast changes are applied as data augmentation\footnote{Our code is available at \url{https://github.com/voyazici/orderless-rnn-classification}}. 

\begin{figure}
\centering
  \includegraphics[width=.9\linewidth,trim={0 0.0cm 0 0.1cm},clip]{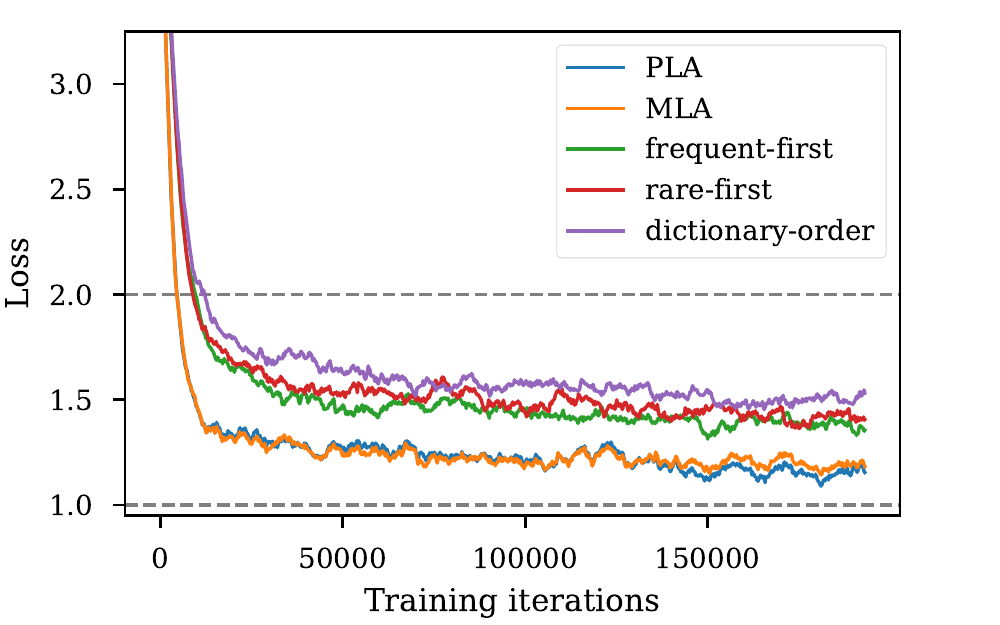}
  \caption{\small Loss curves for different training strategies of CNN-RNN models on the MS-COCO. The graph clearly shows that our strategies, MLA and PLA, obtain significantly lower losses. PLA obtains a slightly better loss from 140,000 iterations onward. This is also reflected in the better performance of PLA for multi-label classification.}
  \label{fig:losses}
\vspace{-1.6em}
\end{figure}

\begin{table}[]
\caption{\small Ratio of duplicates and order-rigidness of different ordering methods on the MS-COCO validation dataset. Results show that our methods do not produce any duplicates and manage to produce label predictions with varying orders (as measured by the order-rigidness).}
\begin{tabular}{l|ll}
\hline
Algorithms  & Ratio of duplicates & Order-rigidness \\ \hline
Random order & 57.86\%               & 67.00\%        \\
Freq. first & 23.84\%              & 100.00\%        \\
Rare-first  & 29.61\%              & 100.00\%        \\
Dict. order & 32.90\%              & 100.00\%        \\ \hline
MLA         & 0.10\%              & 82.87\%        \\ 
PLA        & 0.04\%               & 80.25\%        \\ \hline
\end{tabular}
\label{tab:analysis}
\end{table}

\subsection{Comparison ordering approaches and analysis}
We first compare our method to other CNN-RNN optimization strategies such as frequent-first and rare-first, and evaluate several properties of the different methods. Next we compare it against the state-of-the-art on the MS-COCO, NUS-WIDE, WIDER Attribute and PA-100K datasets.

\minisection{Evaluating training strategies}
First, we compare the different strategies to train CNN-RNN architectures presented in literature: frequent-first~\cite{wang2016cnn}, rare-first~\cite{jin2016annotation} and dictionary-order~\cite{jin2016annotation}.

As can be seen in Figure~\ref{fig:losses}, our proposed strategies MLA and PLA, which dynamically align the ground truth labels with the predicted labels, train faster and obtain a lower overall loss. The rare and frequent-first approaches obtain substantially higher losses. A significant part of the difference between our approaches and these baselines is that they could potentially obtain a non-zero loss on images in which the model perfectly predicts the correct classes but in the wrong order, as can be seen in Figure~\ref{fig:loss_vis}. 
For these images, the backpropagated gradient will try to force the prediction to be in the predefined order (a wasted effort in terms of improving the accuracy) despite the sub-optimality of such order for the particular image in some cases, like when the object that should be predicted first is much smaller than the other objects (see Figure~\ref{fig:freq_rare_problems}).

Next, we analyze the number of duplicate labels generated by the various learning strategies (see Table~\ref{tab:analysis}). To provide a baseline reference, we also include \emph{random order} in the table, which refers to a setting where during training the order of the ground truth labels is randomly selected for each mini-batch. The results show that our method manages to learn not to repeat labels that have already been predicted for an image. In principle, one might think, this should be easy for an LSTM to learn. However, because of the imposed order for the frequent and rare-first approaches, and the resulting confusing backpropagated gradients, the LSTM does not learn this and produces many duplicates. Note that duplicates are not penalizing the overall accuracy of the system, since we remove them in a post-processing step. We would also like to point out here that Chen et al.~\cite{chen2018order} require an explicit module for the removal of duplicates generated by their approach, while we train a model that does not generate duplicates in the first place.

In Table~\ref{tab:analysis}, we show the results for order-rigidness. They show that the methods which impose a fixed order are actually always predicting labels in that order, as indicated by a 100.00\% score. Our methods obtain a 80.25\% and a 82.87\% score, showing that we have no fixed order for the labels and that it is dynamically adjusted to the image.

In Table~\ref{tab:propagation_times}, we show the efficiency of the proposed methods. They require one forward pass and then we apply the Hungarian algorithm to align LSTM predictions and labels (see Eqs.~\ref{eq:MLA} and \ref{eq:PLA}). PLA alignment is faster because it applies the algorithm only to the wrongly predicted labels.

\subsection{Experimental results}
\begin{table}[]
\centering
\caption{\small Results of different ordering methods on MS-COCO.}
\scalebox{0.85}{
\begin{tabular}{l|lll|lll}
\hline
Algorithms   & C-P            & C-R            & C-F1           & O-P   & O-R            & O-F1           \\ \hline
BCE~\cite{wang2016cnn}          & 59.30          & 58.60          & 58.90          & 61.70 & 65.00          & 63.30          \\
BCE          & 68.05          & 59.15          & 63.29          & 72.20 & 65.77          & 68.84          \\
Freq. first  & 70.27          & 56.49          & 62.63          & 72.15 & 64.53          & 68.13          \\
Rare-first   & 65.68          & 61.32          & 63.43          & 70.82 & 64.73          & 67.64          \\
Dict. order  & \textbf{70.98} & 55.86          & 62.52          & 74.14 & 62.35          & 67.74          \\ \hline
MLA          & 68.37          & 60.39          & 64.13          & 72.16 & 66.71          & 69.33          \\
PLA          & 68.66          & 60.54          & 64.34          & 72.67 & 66.89          & 69.66          \\
PLA (atten.) & 70.18          & \textbf{61.96} & \textbf{65.81} & 73.75 & \textbf{67.74} & \textbf{70.62} \\ \hline
\end{tabular}}
\label{tab:order_coco}
\end{table}

\begin{table}[]
\caption{\small Comparison of different ordering methods with average computation times per-image on MS-COCO for ResNet-101.}
\scalebox{0.85}{
\begin{tabular}{l|c|c|c|c}
\multicolumn{1}{c|}{} & \multicolumn{3}{c|}{Training}                                    & Test                     \\ \cline{2-5} 
                      & Forward                  & Alignment & Backward                  &                          \\ \hline
Fixed Order           & \multirow{3}{*}{6.50 ms} & 0 ms      & \multirow{3}{*}{14.80 ms} & \multirow{3}{*}{5.90 ms} \\ \cline{1-1} \cline{3-3}
PLA                   &                          & 0.80 ms (Eq.~\ref{eq:MLA})   &                           &                          \\ \cline{1-1} \cline{3-3}
MLA                   &                          & 0.25 ms (Eq.~\ref{eq:PLA})  &                           &                         
\end{tabular}}
\label{tab:propagation_times}
\vspace{-0.3cm}
\end{table}

\minisection{Comparison of different ordering methods} The results of different ordering algorithms can be seen in Table~\ref{tab:order_coco}. All of the models (except BCE), have the attention module described in the previous section, unless stated otherwise. We observed that the BCE models that we train yield much higher results than the ones cited in previous works, which were originally reported by \cite{wang2016cnn} and \cite{Hu_2016_CVPR} for the MS-COCO and NUS-WIDE respectively. Although it is not very clear from~\cite{wang2016cnn}, we think that the difference between our model and theirs is that, during the training, they freeze all the layers except the last one, since when we impose the same restriction to our model, we obtain similar results. Instead, when we allow for full training of the image encoder the results improve significantly, as reported in Tables~\ref{tab:order_coco}. Interestingly, the fully trained BCE models obtain results similar to those of the CNN-RNN models with the same CNN module when they are trained using the rare-first or frequent-first strategies. We would like to indicate that the results reported in Table~\ref{tab:order_coco} are lower than the results that are in the other tables, since we do not exploit augmentations, train fewer epochs and use a part of training set as validation to tune the hyperparameters in these experiments.

\begin{figure}[tb]
  \includegraphics[width=\linewidth]{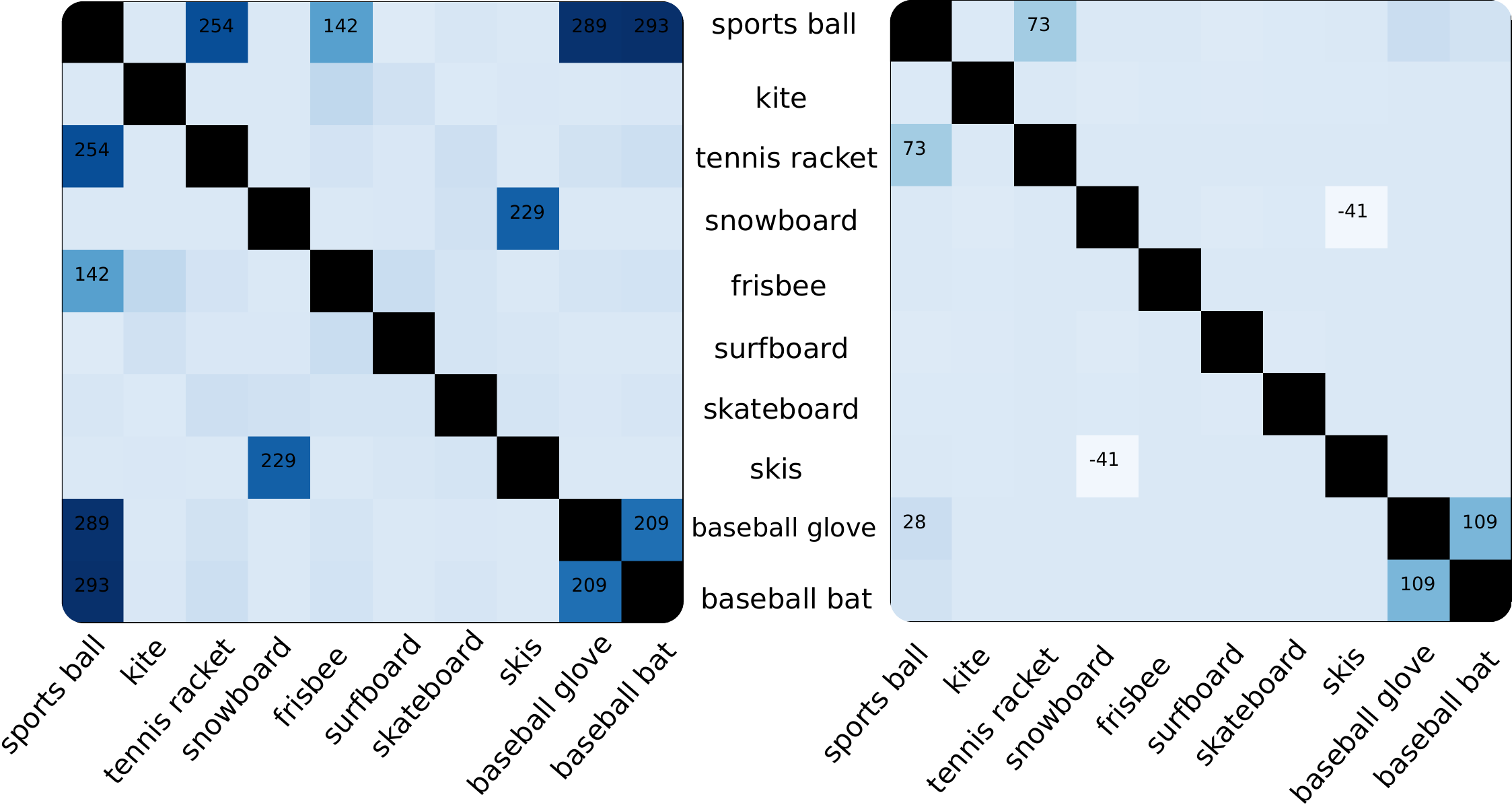}
  \caption{\small Difference of BCE (left) and PLA model (right) co-occurrence matrices with ground truth co-occurrence matrix on the sports super-category of MS-COCO.}
  \label{fig:cooccurrence}
\vspace{-0.1cm}
\end{figure}
When we compare the various strategies of alignment, we see that both our methods, MLA and PLA, clearly outperform the other strategies. Among the other approaches frequent-first yields the best result, although rare-first gives better results with the \textit{per-class} metric since it assigns more weight to the less common classes. The superior performance of PLA with respect to MLA is interesting: we have empirically found that it is better to align with the actual predictions of the network than to align to obtain the minimal loss, as done in MLA. This phenomenon can be observed in Figure~\ref{fig:losses}. Although MLA yields lower losses than PLA in the beginning, PLA's alignment with actual predictions of the network leads to a final lower loss and better accuracies. We think that so many penalizations of correctly predicted labels (see Figure~\ref{fig:cost_matrix}) takes the optimization further away from the global minimum. In fact, the penalization rate during training is as high as 8\% for some classes (e.g.\ baseball bat). For this reason, we select PLA as our default method instead of MLA when we compare our results with the SOTA. Finally, the attention model improves results with a significant gain.

To further investigate the advantages of the LSTM method over BCE, we compare the co-occurrence matrices of the PLA and BCE method with the co-occurrence matrix of the ground truth for the test set. The co-occurrence matrix is computed with $\sum_{i} l^T_i l_i$, where $l_i$ are the ground truth labels for image $i$, and respectively replacing the ground truth labels by the predicted labels $\hat{l}_i$. Next, the co-occurrence matrix of the ground truth is subtracted to that of the predicted labels (the diagonals are ignored, since the co-occurrences of elements with themselves are irrelevant for our analysis). Figure~\ref{fig:cooccurrence} presents the co-occurrence matrices of the BCE and LSTM (PLA) models on the sports supercategory of the MS-COCO dataset. We can observe that BCE has higher co-occurence values, and larger differences with the ground truth. Given that BCE predicts all labels independently from each other, it cannot prevent re-using evidence already used by another prediction. This can be seen, for example, in the many extra co-occurences of predictions for skis and snowboards, as well as for sports ball and frisbee. Similar cases can be observed on the co-occurrence matrices of other supercategories which are available in the supplementary material. Recurrent models, instead, naturally factor in previous predictions at every time step, which leads to a more realistic co-occurrence in the predictions.

\begin{table}[]
\centering
\caption{\small Comparison with state-of-the-art on MS-COCO.}
\scalebox{0.62}{
\begin{tabular}{l|l|lll|lll}
\hline
Algorithms           & Architectures  & C-P & C-R & C-F1  & O-P & O-R & O-F1  \\ \hline
CNN-RNN \cite{wang2016cnn} & VGG16 & 66.00 & 55.60 & 60.40 & 69.20 & 66.40 & 67.80\\
Chen et al.~\cite{chen2018order}   & ResNet152 & 71.60 & 54.80 & 62.10 & 74.20 & 62.20 & 67.70 \\
SR CNN-RNN \cite{liu2017semantic}  & VGG16& 67.40 & 59.83 & 63.39 & 76.63 & 68.73 & 72.47 \\ 
Chen et al.~\cite{chen2018recurrent} & VGG16 & 78.80 & 57.20 & 66.20 & \textbf{84.00} & 61.60 & 71.10 \\
Li et al.~\cite{li2018attentive}  & VGG16 & 71.90 & 59.60 & 65.20 & 74.30 & 69.70 & 71.80  \\ \hline
MS-CNN+LQP \cite{niu2019multi} & ResNet101 & 67.48 & 60.93 & 64.04 & 70.22 & 67.93 & 69.06 \\
LSEP \cite{li2017improving} & VGG16 & 73.50 & 56.40 & 63.82 & 76.30 & 61.80 & 68.29 \\
MLIC-KD-WSD \cite{liu2018mlickdwsd} & VGG16 & - & - & 69.20 & - & - & 74.00  \\
SRN \cite{zhu2017learning} & ResNet101 & 81.60 & 65.40 & 71.20 & 82.70 & 69.90 & 75.80 \\ 
ACfs \cite{guo2019visual} & ResNet101 & 77.40 & 68.30 & 72.20 & 79.80 & 73.10 & 76.30 \\ \hline
PLA & VGG 16 & 73.72 & 63.18 & 68.05 & 78.25 & 68.76 & 73.20 \\ 
PLA & ResNet101 & \textbf{80.38} & 68.85 & \textbf{74.17} & 81.46 & 73.26 & \textbf{77.14} \\ 
PLA & ResNet152 \footnotemark & 75.32 & \textbf{69.62} & 72.36 & 76.85 & \textbf{73.97} & 75.38 \\ \hline
\end{tabular}}
\label{tab:sota_coc}
\end{table}
\footnotetext{Input size of $224\times224$ (smaller than ResNet101) to compare with~\cite{chen2018order}.}

\begin{table}[]
\centering
\caption{\small Comparison with state-of-the-art on NUS-WIDE.}
\scalebox{0.75}{
\begin{tabular}{l|lll|lll}
\hline
Algorithms             & C-P & C-R & C-F1  & O-P & O-R & O-F1  \\ \hline
CNN-RNN \cite{wang2016cnn} & 40.50 & 30.40 & 34.70 & 49.90 & 61.70 & 55.20\\ 
Chen et al.~\cite{chen2018order}   & 59.40 & 50.70 & 54.70 & 69.00 & 71.40 & 70.20 \\
SR CNN-RNN \cite{liu2017semantic}  & 55.65 & 50.17 & 52.77 & 70.57 & 71.35 & 70.96 \\
Li et al.~\cite{li2018attentive}  & 44.20 & 49.30 & 46.60 & 53.90 & 68.70 & 60.40  \\ \hline
LSEP \cite{li2017improving} & \textbf{66.70} & 45.90 & 54.38 & \textbf{76.80} & 65.70 & 70.82 \\
MLIC-KD-WSD \cite{liu2018mlickdwsd} & - & - & \textbf{58.70} & - & - & \textbf{73.70}  \\  \hline
PLA & 60.67 & \textbf{52.40} & 56.23 & 71.96 & \textbf{72.79} & 72.37 \\\hline
\end{tabular}}
\label{tab:sota_nus}
\vspace{-0.2cm}
\end{table}

\minisection{Comparison to state-of-the-art}
We compare our results with several models, grouped into two categories: models that use a CNN-RNN jointly and models that use alternative approaches. CNN-RNN~\cite{wang2016cnn}, SR CNN-RNN~\cite{liu2017semantic} and Chen et al.~\cite{chen2018order} are directly related to our model (see sec.~\ref{sec:related_work} for details).
Also in this category, Chen et al.~\cite{chen2018recurrent} use an LSTM to predict the next region to attend according to the hidden state and the current region, after which they fuse the predictions of each time step. 
Similarly, Li et al.~\cite{li2018attentive} use a recurrent network to highlight image regions to attend, but then employ reinforcement learning to select which regions should be used for the actual prediction. Among the alternative approaches, MS-CNN+LQP~\cite{niu2019multi} tries to explicitly predict the number of tags in images, LSEP~\cite{li2017improving} uses a pairwise ranking approach for training a CNN, and MLIC-KD-WSD~\cite{liu2018mlickdwsd} use knowledge distillation from a teacher network which is trained on a weakly supervised detection task. ACfs~\cite{guo2019visual}, proposes a two-branch network with an original image and its transformed image as inputs and imposes an additional loss to ensure the consistency between attention heatmaps of both versions. SRN~\cite{zhu2017learning} proposes a Spatial Regularization Network that generates attention maps for all labels and models the label correlations via learnable convolutions. HP-Net~\cite{liu2017hydraplus} proposes a novel attention module to train multi-level and multi-scale attention-strengthened features for pedestrian analysis. The results on the mentioned datasets can be seen in Tables~\ref{tab:sota_coc},~\ref{tab:sota_nus},~\ref{tab:sota_wider} and~\ref{tab:sota_pa100k}.\footnote{Ge et al.~\cite{ge2018multi} also evaluate  on COCO, however they require additional semantic maps which makes their model incomparable.}

\begin{table}[]
\centering
\caption{\small Comparison with state-of-the-art on WIDER Attribute.}
\scalebox{0.75}{
\begin{tabular}{l|lll|lll}
\hline
Algorithms             & C-P & C-R & C-F1  & O-P & O-R & O-F1  \\ \hline
SRN \cite{zhu2017learning} & - & - & 75.90 & - & - & 81.30 \\
ACfs \cite{guo2019visual} & 81.30 & 74.80 & 77.60 & 84.10 & 80.70 & 82.40 \\
PLA & \textbf{81.69} & \textbf{75.87} & \textbf{78.67} & \textbf{84.99} & \textbf{81.36} & \textbf{83.13} \\ \hline
\end{tabular}}
\label{tab:sota_wider}
\end{table}

\begin{table}[]
\centering
\caption{\small Comparison with state-of-the-art on PA-100K.}
\scalebox{0.75}{
\begin{tabular}{l|lll|lll}
\hline
Algorithms             & Precision & Recall & F1  & Accuracy  \\ \hline
DM \cite{li2015multi} & 82.24 & 80.42 & 81.32 & 70.39 \\
HP-Net \cite{liu2017hydraplus} & 82.97 & 82.09 & 82.53 & 72.19 \\
ACfs \cite{guo2019visual} & \textbf{88.97} & 86.26 & \textbf{87.59} & 79.44 \\
PLA & 88.50 & \textbf{86.67} & 87.58 & \textbf{79.83} \\ \hline
\end{tabular}}
\label{tab:sota_pa100k}
\vspace{-0.3cm}
\end{table}

On the MS-COCO dataset, we get higher F1 scores than all other CNN-RNN models. MLIC-KD-WSD~\cite{liu2018mlickdwsd} also achieves notable results by exploiting knowledge distillation from a teacher network. Only for MS-COCO, we resize input images to $288 \times 288$ to be able to compare PLA ResNet-101 model with the ACfs~\cite{guo2019visual}. We also show the results for the ResNet-152 architecture to compare with~\cite{chen2018order}.

On the NUS-WIDE dataset, we surpass all other CNN-RNN models by a significant margin. Our results are especially remarkable for per-class F1 score, a more relevant metric for unbalanced datasets such as this. The globally best results are achieved in terms of overall and per-class F1 score by MLIC-KD-WSD~\cite{liu2018mlickdwsd}. All the models that are compared on NUS-WIDE dataset use the VGG16 as the backbone network. We do not display the results of \cite{niu2019multi} since they use a different split of the dataset.

To be comparable with other models, we use ResNet-101 and ResNet-50 architectures for WIDER Attribute and PA-100K respectively. These two datasets have human attributes that are related to gender, appearance, clothing, etc. as labels. Therefore, label correlations are not common among these two datasets, which is a drawback for CNN-RNN models. In spite of that, our CNN-RNN model manages to surpass the other models.
\vspace{-0.15cm}
\section{Conclusions}
\vspace{-0.1cm}
We proposed an approach for training orderless LSTM models applied to multi-label classification task. Previous methods imposed an ordering on the labels to train the LSTM model, Typically frequent-first or rare-first orderings were used. Instead, we proposed two alternative losses  which dynamically order the labels based on the prediction of the LSTM model.  Our approach is unique in that seldomly generates any duplicate prediction, and that it minimizes the loss faster than the other methods. Results show that a standard CNN-RNN architecture, when combined with our proposed orderless loss, obtains the state-of-the-art results for multi-label classification on several datasets.

\minisection{Acknowledgements.} We thank the Spanish project TIN2016-
79717-R, the Industrial Doctorate Grant 2016 DI 039 of the
Ministry of Economy and Knowledge of the Generalitat de Catalunya, and also its CERCA Program.

{\small
\bibliographystyle{ieee_fullname}
\bibliography{shortstrings,egbib}
}

\newpage

\begin{center}
\textbf{\LARGE Supplementary Material for Orderless Recurrent Models for Multi-label Classification}
\end{center}

\vspace{1.3cm}
\setcounter{section}{0}
\setcounter{figure}{0}

\section{Introduction}
This is the supplementary material of the paper \emph{Orderless Recurrent Models for Multi-label Classification}. We first exhibit some images whose labels are sorted differently by the proposed \emph{predicted label alignment} (PLA) and frequent-first approach. Then, we show co-occurrence matrices of different super-categories of the MS-COCO dataset computed by the best BCE and LSTM models.

\section{Qualitative comparison of PLA and other methods}
\begin{figure}
    \includegraphics[width=\linewidth]{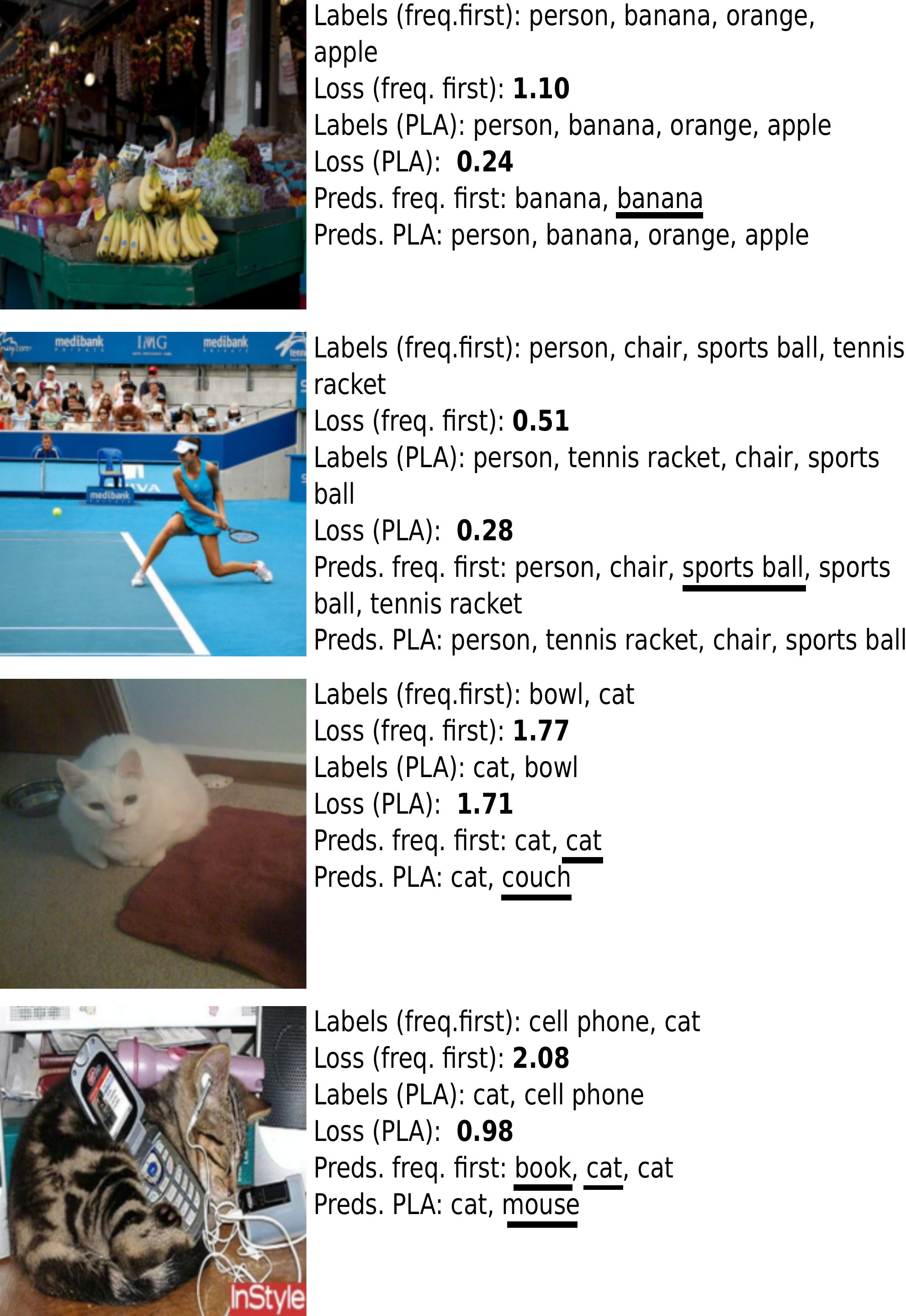}
    \caption{Comparisons of orders yielded by the PLA and frequent-first approaches (wrong or duplicate predictions are underlined).}
    \label{fig:first_page}
\end{figure}

\begin{figure*}
    \includegraphics[width=\linewidth]{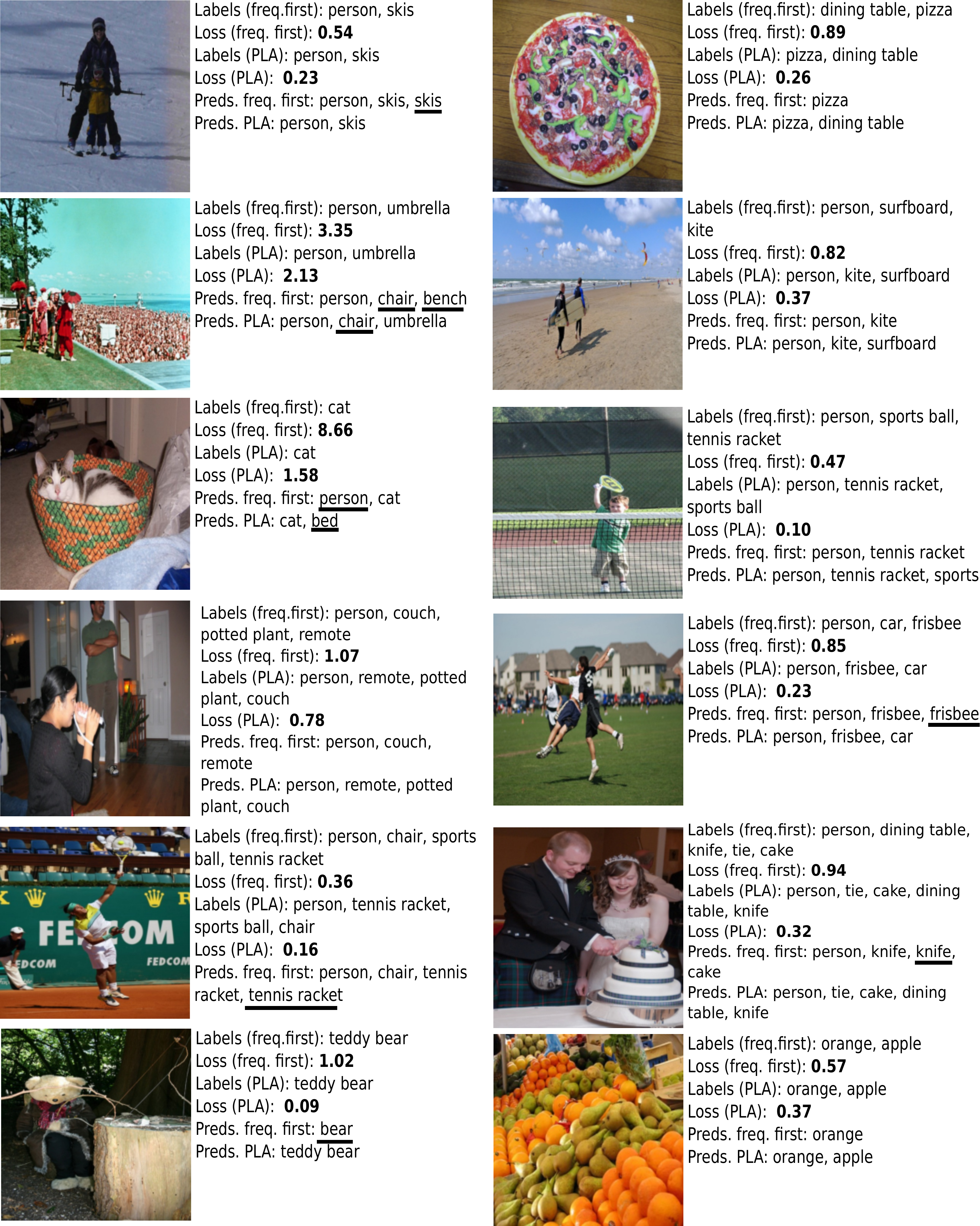}
    \caption{See caption of Figure~\ref{fig:first_page}.}
    \label{fig:second_page}
\end{figure*}

In Figures~\ref{fig:first_page}-\ref{fig:second_page}, we can see different orders yielded by the PLA and frequent-first approaches. The images are chosen to emphasize the problems with the approaches that use predefined orders. As can be seen in the images, the frequent-first approach always predicts the labels in the same order. This leads to confusion in case of dominant but less frequent objects or minor but more frequent objects in an image. Then, this confusion leads to duplicate predictions in different time steps.

\section{Additional co-occurrence matrices of LSTM and BCE models}

\begin{figure*}
    \centering
    \includegraphics[width=0.8\linewidth]{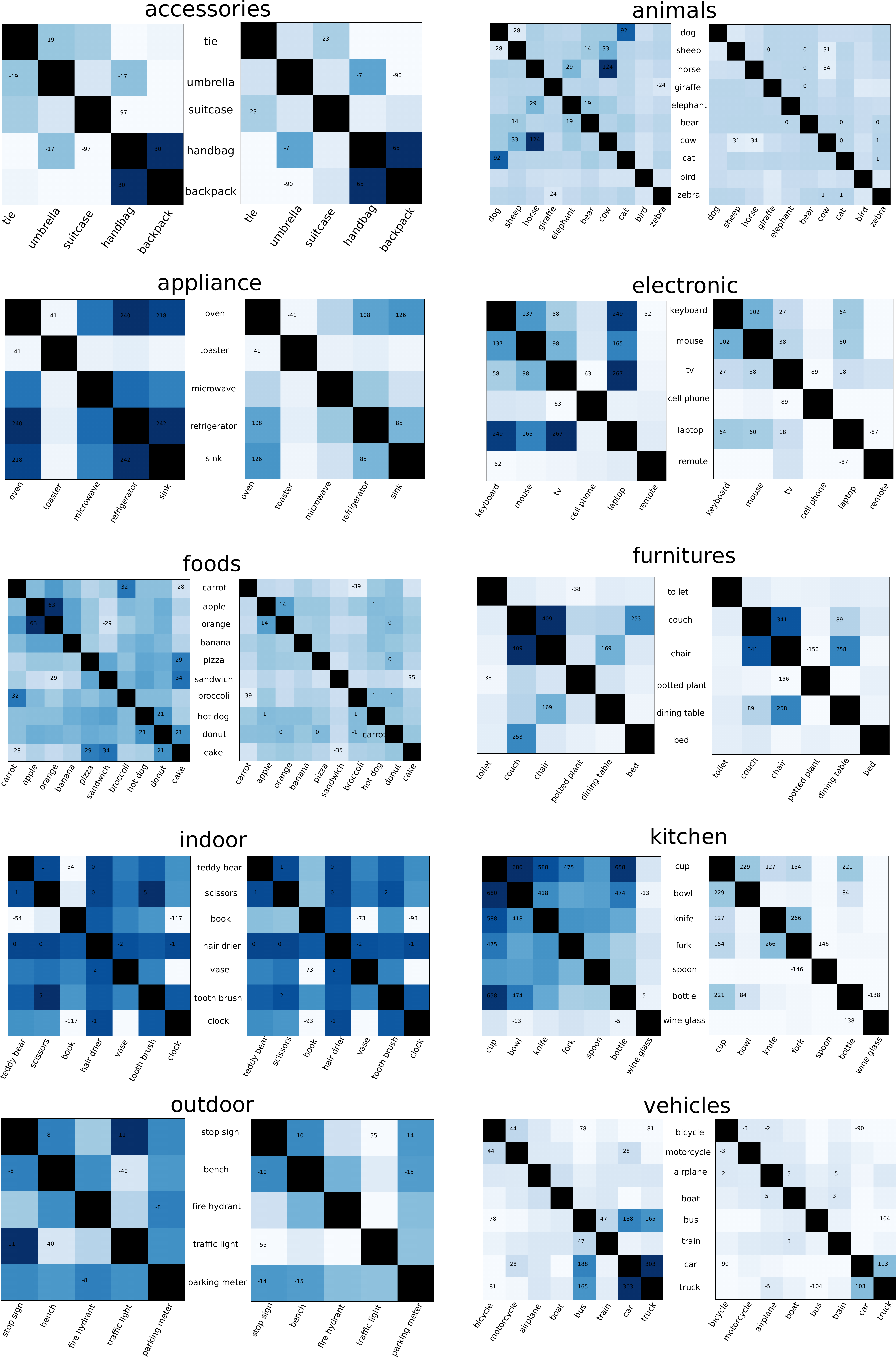}
    \caption{Co-occurence matrices for BCE (left) and PLA (right) models. BCE re-uses evidence to predict different objects, and hence has higher co-occurence levels due to false positives.}\label{fig:all_comatrices}
\end{figure*}

In Figure~\ref{fig:all_comatrices}, the predicted class co-occurence matrices for binary cross entropy (BCE) and predicted label alignment (PLA) models can be seen. The levels of co-occurence in BCE are noticeably higher than those on PLA, as it re-uses the same parts of image for different predictions of similar objects (e.g. bike and motorbike). This can be observed especially on the animals, food, vehicle and kitchen super-categories. In the animals super-category the BCE model overshoots co-occurrence of dogs-cats and horses-cows, while in the vehicles the confusion is on the buses, trucks and cars. In the kitchen super-category the confusion is the worst since most of the images are images of entire kitchens and the BCE model uses the entire scene for different predictions. On the other hand, the LSTM model has much lower differences with the ground truth, since the previous predictions are taken into account at every time step.

\end{document}